\def\aaaianonymous{true}
\documentclass[letterpaper]{article} 

\ifdefined\aaaianonymous
    \usepackage{aaai2026}              
\fi

\usepackage{appendix} 
\usepackage{times}  
\usepackage{helvet}  
\usepackage{courier}  
\usepackage[hyphens]{url}  
\usepackage{graphicx} 
\usepackage{natbib}  
\usepackage{caption} 
\usepackage{amsmath}    
\usepackage{booktabs}   
\usepackage{tabularx}   
\usepackage{array}          
\newcolumntype{C}[1]{>{\centering\arraybackslash}m{#1}} 
\newcolumntype{Y}{>{\centering\arraybackslash}X}        
\usepackage{placeins}  
\usepackage{booktabs}  
\usepackage{tabularx}  
\usepackage{multirow}   
\usepackage{siunitx}    
\usepackage{subcaption}  
\newcolumntype{M}[1]{>{\raggedright\arraybackslash}m{#1}}
\usepackage{amsmath} 
\usepackage{mathrsfs}
\usepackage{tabularx,array}                     
\newcolumntype{L}[1]{>{\raggedright\arraybackslash}m{#1}}

\usepackage{algorithm}
\usepackage{algorithmic}

\usepackage{newfloat}
\usepackage{listings}
\DeclareCaptionStyle{ruled}{labelfont=normalfont,labelsep=colon,strut=off} 
\floatstyle{ruled}
\newfloat{listing}{tb}{lst}{}
\floatname{listing}{Listing}

\ifdefined\aaaianonymous
    \title{HISE-KT: Synergizing Heterogeneous Information Networks and LLMs \\ 
    for Explainable Knowledge Tracing with Meta-Path Optimization}
\else
    \title{HISE-KT: Synergizing Heterogeneous Information Networks and LLMs \\ 
           for Explainable Knowledge Tracing with Meta-Path Optimization}
\fi

\author{
    Zhiyi Duan\textsuperscript{\rm 1},
    Zixing Shi\textsuperscript{\rm 1},
    Hongyu Yuan\textsuperscript{\rm 1},
    Qi Wang\textsuperscript{\rm 2}\thanks{Corresponding author.}
}

\affiliations {
    \textsuperscript{\rm 1}Department of Computer Science, Inner Mongolia University, Hohhot, Inner Mongolia\\
    \textsuperscript{\rm 2}School of Artificial Intelligence, Jilin University, Changchun, Jilin\\
    duanzy@imu.edu.cn, zixingshi@mail.imu.edu.cn, yuanhongyu\_1997@163.com, qiwang@jlu.edu.cn
}

\usepackage{bibentry}

\begin{document}

\maketitle

\begin{abstract}
Knowledge Tracing (KT) aims to mine students’ evolving knowledge states and predict their future question-answering performance. Existing methods based on heterogeneous information networks (HINs)  are prone to introducing noises due to manual or random selection of meta-paths and lack necessary quality assessment of meta-path instances. Conversely, recent large language models (LLMs)-based methods ignore the rich information across students, and both paradigms struggle to deliver consistently accurate and evidence-based explanations. To address these issues, we propose an innovative framework, \textit{HIN-LLM Synergistic Enhanced Knowledge Tracing} \textit{(HISE-KT)}, which seamlessly integrates HINs with LLMs. HISE-KT first builds a multi-relationship HIN containing diverse node types to capture the structural relations through multiple meta-paths. The LLM is then employed to intelligently score and filter meta-path instances and retain high-quality paths, pioneering automated meta-path quality assessment. Inspired by educational psychology principles, a similar student retrieval mechanism based on meta-paths is designed to provide a more valuable context for prediction. Finally, HISE-KT uses a structured prompt to integrate the target student's history with the retrieved similar trajectories, enabling the LLM to generate not only accurate predictions but also evidence-backed, explainable analysis reports. Experiments on four public datasets show that HISE-KT outperforms existing KT baselines in both prediction performance and interpretability.
\end{abstract}

\section{Introduction}

\par Knowledge Tracing (KT), as a core task of educational data mining and intelligent education systems, aims to dynamically model the knowledge state of students based on their historical interaction sequences and to predict their performance on future learning content \cite{corbett1994knowledge}. Accurate KT models are critical for personalized learning-path recommendation, adaptive learning-resource allocation, and precise instructional interventions, serving as key technological support for improving educational quality and efficiency \cite{shen2024survey, duan2024towards}. 

\par Recently, researchers have attempted to leverage heterogeneous information networks (HINs) to enhance both the performance and interpretability of KT models \cite{shi2018heterogeneous, xu2023improving}. A HIN can naturally integrate various types and levels of entities in educational scenarios (e.g., students, questions, and knowledge concepts) and their rich interaction relationships (e.g., answering, inclusion, and assessment), providing a structured representation framework for modeling complex learning processes \cite{sun2024question}. By defining meta-paths to characterize specific semantic relationship patterns among entities, HINs can effectively capture the implicit and complex associations in learning processes \cite{sun2011pathsim}. However, existing HIN-based KT methods face significant challenges. First, meta-path selection often relies on expert knowledge or random walks, inevitably producing redundant or low-information path instances. These instances increase computational overhead and introduce noise, ultimately degrading model performance \cite{meng2015discovering, liu2023heteedgewalk}. Second, model interpretability is typically limited to the network-structure level (e.g., path weights), making it difficult to provide high-level explanations that are semantically clear and aligned with human cognition. These issues limit the credibility and applicability of such models in practical educational decision-making.

\par Meanwhile, large language models (LLMs) have demonstrated remarkable performance across various tasks due to their powerful capabilities in semantic understanding, contextual reasoning, and natural language generation \cite{zhao2023survey, achiam2023gpt, zhou2024self, guo2025deepseek}. Preliminary studies applying LLMs to KT have shown encouraging progress. LLMs possess a natural advantage in generating explanations in natural language, which may enhance the interpretability of KT models. However, existing LLM-based KT methods primarily rely on instruction fine-tuning or prompt engineering over answering sequences \cite{li2025cikt, jung2024clst}, failing to structurally model the complex higher-order interactions among students, questions, and knowledge concepts (e.g., indirect associations formed via mediating nodes such as students' abilities and question difficulties). This limitation hampers the capture of deeper learning patterns, constrains performance gains, and induces behavioral explanation hallucinations due to evidence scarcity.

\par To address these limitations, we propose an innovative \textit{HIN-LLM Synergistic Enhanced Knowledge Tracing (HISE-KT)} framework, which integrates the structured-relationship strengths of HINs with the semantic understanding, reasoning, and natural-language generation capabilities of LLMs to improve both performance and interpretability. Specifically, to represent the complex interaction relationships in the learning process, we first construct a multi-relationship HIN containing multiple node types and multi-dimensional semantic relationships. Then, an LLM-based scoring method is proposed to perform multi-dimensional semantic evaluation of the generated meta-path instances, mitigating the randomness and redundancy of traditional meta-path selection. The Top-$K$ meta-paths with the highest discriminative power and information content are retained to construct an initial candidate student retrieval set. 
Relative Evaluation Theory argues that assessing a student against similar peers filters out common noise and highlights genuine ability differences \cite{gibbons1990relative}. Social Comparison Theory further posits that individuals gauge their own competence by comparing themselves to those with comparable traits, yielding the most meaningful benchmarks \cite{suls2012social}. Guided by these, a rule-based collaborative filtering that considers path matching, knowledge state similarity, and historical performance trends is proposed to select the Top-$S$ most relevant similar students. The historical trajectories of these students supply valuable, comparable context for the target learner. Building on this context, a structured prompt is designed to enable zero-shot KT prediction and generate an explainable analysis report, which clarifies the prediction basis, highlights potential learning difficulties, and offers actionable improvement suggestions, thereby enhancing decision transparency. \textbf{The major contributions  are summarized as follows}:
\begin{itemize}
    \item We propose HISE-KT, the first KT framework that synergizes structured relational modeling with HINs and generative semantic intelligence with LLMs, establishing a new paradigm for KT.
    \item We establish an LLM-powered meta-path optimization mechanism, where semantic scoring and selective filtering automatically eliminate noisy paths while preserving  meaningful interactions, overcoming redundancy and inefficiency issues in HIN-based KT.
    \item We develop a rule-based collaborative filtering mechanism inspired by Relative Evaluation and Social Comparison Theory, providing contextual anchors that mitigate LLM hallucinations and enable traceable explanations.
    \item Experiments on multiple publicly available educational datasets verify that HISE-KT outperforms state-of-the-art baselines in both predictive performance and explanation quality.
\end{itemize}

\section{Related Work}

\subsection{HIN-Based KT Methods}
Recent studies have tried to use HINs to model the relationships in the learning process. STHKT \cite{li2025sthkt} constructs heterogeneous graphs and combines topological Hawkes processes with graph convolutional networks to fuse spatiotemporal information. MGEKT \cite{qiu2024knowledge} uses meta-paths to capture higher-order semantics in a heterogeneous graph, while integrating a gated attention mechanism to model both student interactions and long-term dependencies. SimQE \cite{sun2024question} captures the similarity of questions through biased random walks of meta-paths on a weighted HIN to enhance question representation. However, these methods model entities in KT relatively simplistically and do not evaluate the quality of meta-path instances, resulting in redundant samples that hinder the modeling of truly discriminative high-order relationships.

\subsection{LLM-Based KT Methods}
LLM is introduced into KT to improve interpretability due to its excellent semantic understanding and generation capabilities. Some works employ instruction fine-tuning of LLMs to directly learn from student interaction data (e.g.,  CLST \cite{jung2024clst}, LLM-KT \cite{wang2025llm}, CIKT \cite{li2025cikt}). LLM-KT performs instruction fine-tuning on the LLMs via Low-Rank Adaptation, encapsulating student response history and question information into structured prompts while integrating sequential behavior embeddings with textual context to enhance knowledge tracing performance \cite{wang2025llm}. Meanwhile, other works achieve KT by constructing structured prompts (e.g., EFKT \cite{li2024explainable}, LOKT \cite{kim2024beyond}). EFKT utilizes few-shot prompting to leverage the reasoning and generation capabilities of LLMs for completing KT with limited practice records and generating interpretable natural language prediction reports \cite{li2024explainable}. However, these methods rely solely on a single student's historical interaction data for KT prediction, without considering the cross-student interaction information, which may lead to hallucinations and reduce the reliability of predictions and explanations.


\begin{figure*}[t]
    \centering
    \includegraphics[width=\textwidth]{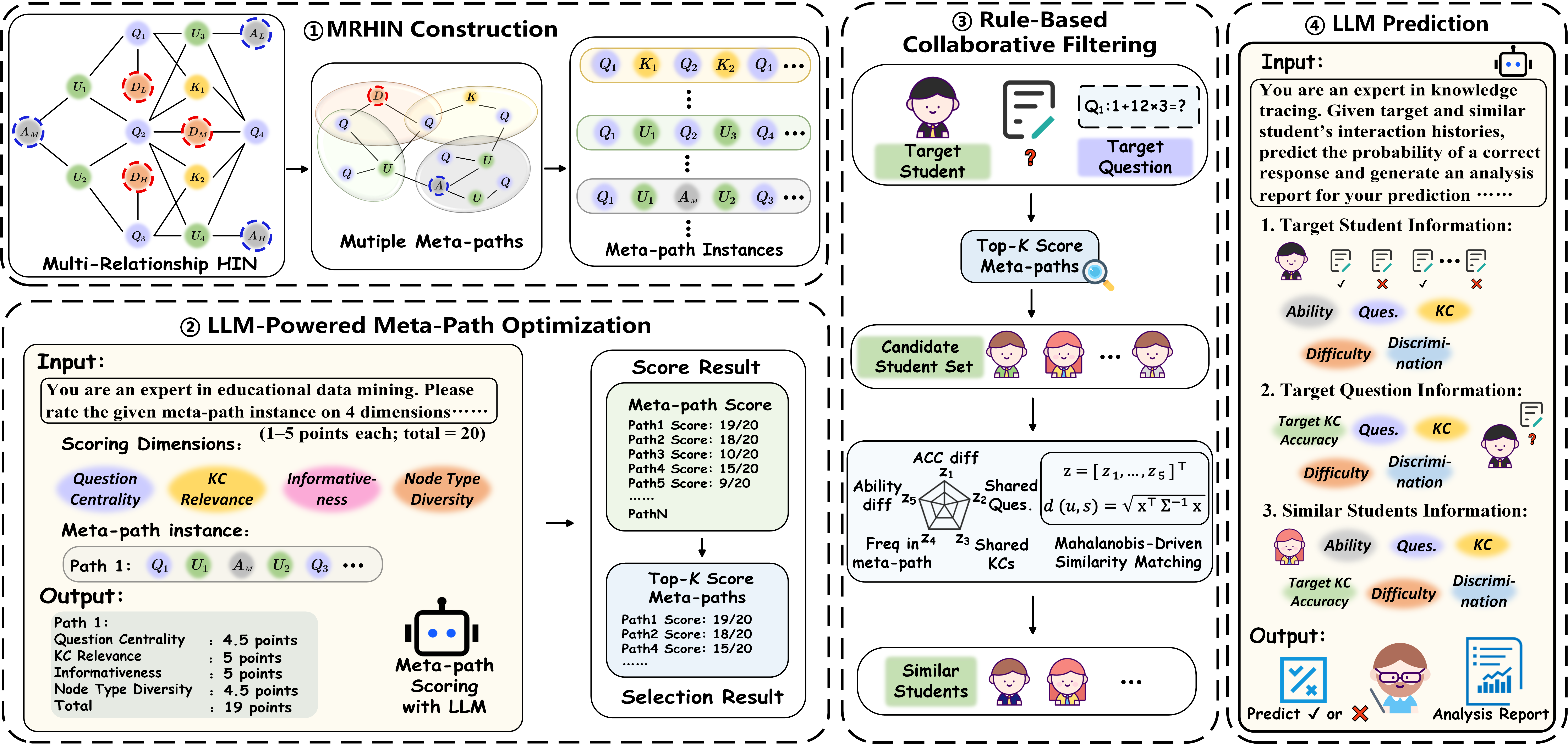}
    
    \caption{The framework of HISE-KT: (1) Build a multi-relationship HIN, then define and instantiate multiple meta-paths; (2) Employ the LLM to score and select high-quality meta-path instances; (3) Retrieve the target student's similar peers through high-quality meta-path; (4) Input cross-student information into LLM for KT prediction and analysis reports generation.}
    
    \label{fig:Framework}
\end{figure*}

\section{Methodology}
In this section, we introduce HISE-KT, the overall framework is shown in Fig.~\ref{fig:Framework}. The framework comprises four sequential modules: (1) MRHIN Construction Module; (2) LLM-Powered Meta-Path Optimization Module; (3) Rule-Based Collaborative Filtering Module; (4) LLM Prediction Module. The following subsections will introduce the specific implementation methods of each module in detail.




\subsection{Multi-Relationship Heterogeneous Information Network Construction Module}
This module aims to construct a Multi-Relationship Heterogeneous Information Network (MRHIN) to better model the structured relationships between students, questions, and knowledge concepts. Existing HINs often only consider the shallow relationships between students, questions, and knowledge concepts, which ignore student attributes and question attributes. Therefore, we construct MRHIN, which contains five types of nodes: student ($U$), question ($Q$), knowledge concept ($K$), student ability ($A$), and question difficulty ($D$). There are four types of edges: question-student ($Q-U$), question-knowledge concept ($Q-K$), question-question difficulty ($Q-D$), and student-student ability ($U-A$). These edges are used to model the deep structured relationships between nodes, forming a multi-relationship heterogeneous information network.

\noindent\textbf{External Node Construction.}
This section employs the two-parameter logistic Item Response Theory (IRT-2PL) model \cite{birnbaum1968some} to estimate each student's latent ability and each question's difficulty. Since raw ability and difficulty parameter values are continuous, we discretize them into three levels (e.g., Low, Medium, High) based on their mean and standard deviation. These levels are then used to construct the external nodes \(A\) and \(D\). The detailed IRT-2PL and level division formulas are provided in Appendix A.







\noindent\textbf{Meta-Path Construction.}
In order to capture the latent high-order relationships among five node types ($U$, $Q$, $K$, $A$, and $D$) in MRHIN, we design 14 meta-paths to capture the relationships between these nodes. Since different meta-paths represent different semantics, these meta-paths are divided into two categories: \emph{basic meta-paths} and \emph{composite meta-paths}. \emph{Basic meta-paths} are defined as those that reflect a single semantic and cannot be decomposed into a second independent semantic. We define four \emph{basic meta-paths}, which capture the most direct relationships between nodes:
\begin{itemize}
    \item \textit{Q-U-Q}: two questions are answered by the same student, capturing the relationship between $Q$ and $U$.
    \item \textit{Q-K-Q}: two questions examine the same knowledge concept, capturing the relationship between $Q$ and $K$.
    \item \textit{Q-D-Q}: two questions belong to the same question difficulty, capturing the relationship between $Q$ and $D$.
    \item \textit{Q-U-A-U-Q}: two questions are answered by students with same ability, capturing the relationship among $Q$, $U$, and $A$.
\end{itemize}
\emph{Composite meta-paths} are composed of basic paths. In order to capture the complex cross-semantic relationships between multiple nodes, we define ten meta-paths. Finally, for each question node, \(N\) path instances are sampled for every meta-path via an equal-probability random walk on MRHIN. The specific definition of the ten \emph{composite meta-paths} is provided in Appendix B.

\subsection{LLM-Powered Meta-path Optimization Module}
Random walk sampling generates meta‑path instances that cover multi‑dimensional semantics for each target question but also introduces redundant and noisy links. We employ an LLM to score each meta-path instance across four dimensions and select the Top‑$K$ highest scoring paths for each meta‑path template. These selected paths better capture the structural relationships in the MRHIN.

Define the instantiated path as \(P=(v_0,\dots ,v_L)\), where \(v_0=q_0\) denotes the target question node, \(Q(P)\) represents the set of question nodes in the path, \(k^\ast\) is the target knowledge concept, and \(\mathrm{dist}(x,y)\) measures the shortest distance between two nodes~\(x\) and~\(y\) in the MRHIN. Uppercase letters indicate node types (\(U,Q,K,A,D\)), while lowercase letters denote concrete nodes (e.g., \(q_0\)). In every formula, “\(\propto\)” indicates a positive correlation between the score and the meta-path instance quality. The formulaic expressions of the four dimensions are shown below, and the complete scoring prompt is provided in Appendix C.

\noindent\textbf{Question Centrality.}
When the number of random walk steps increases, subsequent nodes in the path tend to deviate from the target question, resulting in a lack of direct relevance at the question level. We use \textit{Question Centrality} dimension $C_q$ to measure whether the path consistently maintains a star-like closed structure centered around $q_0$. The closer and more frequently question nodes revisit \(q_0\), the higher the question centrality, we have:

\begin{equation}
C_q(P) \;\propto\;
\left(1-\frac{1}{|Q(P)|}\sum_{q\in Q(P)}\frac{\mathrm{dist}(q_0,q)}{L}
\right)
\end{equation}
where \(\mathrm{dist}(q_0,q)\) is the distance between \(q_0\) and \(q\) in MRHIN, $L$ is the length of the meta-path instance.

\noindent\textbf{Knowledge Concept Relevance.}
High-quality paths should closely revolve around the target knowledge concept $k^*$ to highlight semantic relevance at the knowledge level. If the path frequently turns to irrelevant knowledge concepts, it provides limited or even detrimental predictive performance. The \textit{Knowledge Concept Relevance} dimension $R_{\text{KC}}$ measures the proportion of questions related to $k^*$ in the path; a higher proportion indicates more focused semantic coherence, we have: 
\begin{equation}
R_{\text{KC}}(P) \;\propto\;
\frac{\bigl|\{\,q\in Q(P)\mid k^\ast\in\mathrm{KC}(q)\}\bigr|}{|Q(P)|}
\end{equation}
where $\mathrm{KC}(q)$ is the set of knowledge concepts contained in question $q$.

\noindent\textbf{Informativeness.}
During random walks, paths may cycle among a few nodes, creating redundant links and diluting effective signals. The \textit{Informativeness} dimension $I_{\text{info}}$ encourages paths to continuously introduce new nodes  $(U, Q, K)$ to obtain more independent evidence, we have:

\begin{equation}
I_{\text{info}}(P) \;\propto\;
\frac{\mathrm{distinct}_{\neg q_0,k^\ast}(P)}
{\bigl|\{\,v\in P\mid \text{type}(v)\in\{U,Q,K\}\}\bigr|}
\end{equation}
where $\mathrm{distinct}_{\neg q_0,k^\ast}(P)$ denotes the unique occurrences of three node types $(U, Q, K)$, while ignoring repeated visits to the initial question node $q_0$ and target knowledge concept node $k^*$. A higher proportion of distinct nodes yields a higher informativeness score.

\noindent\textbf{Node Type Diversity.}
The two external nodes, $A$ and $D$, can reveal higher-order interactions between student proficiency and question intensity. A complete and balanced level distribution provides rich information for capturing structured relationships in the MRHIN. The \textit{Node Type Diversity} dimension $D_{\text{type}}$ encourages paths to simultaneously cover various levels of $A$ and $D$ while avoiding extreme imbalances. The more comprehensive the levels and the more balanced their distribution, the higher the node type diversity score, we have:

\begin{equation}
D_{\text{type}}(P) \;\propto\;
-\frac{1}{\log|\mathcal{L}|}\sum_{t\in\mathcal{L}} p_t \log p_t
\end{equation}
where \(\mathcal{L}=\{A_{\text{Low}},A_{\text{Medium}},A_{\text{High}},D_{\text{Low}},D_{\text{Medium}},D_{\text{High}}\}\), \(p_t\) denotes the relative frequency of nodes belonging to Level \(t\).

LLM generates scores for each path of the four dimensions, with a maximum score of 5 points per dimension and a total score of 20 points:

\begin{equation}
S(P)=C_q+R_{\text{KC}}+I_{\text{info}}+D_{\text{type}}
\end{equation}
where $S$ denotes the total score of a meta-path instance.

\subsection{Rule-Based Collaborative Filtering Module}
The high-quality meta-paths selected by LLM can effectively  capture the structural relationships between the five types of nodes $(U, Q, K, A, D)$ in MRHIN. Inspired by Relative Evaluation \cite{gibbons1990relative} and Social Comparison Theory \cite{suls2012social} in educational psychology, we recognize that the performance and ability evaluation of individual learners is often more meaningful and valuable when compared with their similar peers. 

Building on these paths, we propose a rule-based collaborative filtering mechanism to retrieve students with similar potential knowledge states to the target student, thereby capturing cross-student interaction information. This module can be divided into three stages: Candidate Student Set Construction, Cross‑Student Similarity Encoding, and Mahalanobis‑Driven Similarity Matching.

\noindent\textbf{Candidate Student Set Construction.}
Based on the target student's interaction history, the target question to be tested can be obtained. According to all the high-quality meta-path instances retained under the target question, the candidate student set $\mathcal{C}$ is constructed after removing the duplicate student nodes that are appearing in these paths. $\mathcal{C}$ incorporates student attributes and relevant historical interaction information. Due to the constraints of the meta-path quality assessment, the students in $\mathcal{C}$ have potential connections with the target question.

\noindent\textbf{Cross‑Student Similarity Encoding.}
To quantify the difference between the target student \(u\) and a candidate student \(s\), we construct a five-dimensional feature vector as:
\begin{equation}
    \mathbf{z}_{u,s}
          = [\,z_{1},z_{2},z_{3},z_{4},z_{5}\,]^{\top}
\end{equation}
where
\begin{equation}
    z_{1} = \lvert \theta_{u}-\theta_{s} \rvert
\end{equation}
here, $z_{1}$ encodes the difference in student ability $\theta$ between $u$ and $s$. The closer the abilities, the more similar the two students are.

\begin{equation}
    z_{2} = \frac{c}{\lvert K \rvert}
            \sum_{k\in K}
            \bigl\lvert \mathrm{acc}_{u}(k)-\mathrm{acc}_{s}(k) \bigr\rvert  
\end{equation}
 here, $c$ is a scaling constant that controls the decay rate. $z_{2}$ encodes the difference in accuracy between $u$ and $s$ on shared knowledge concepts $K$. A smaller accuracy difference implies greater similarity.

\begin{equation}
    z_{3} = (1+N_{\mathrm{Q}})^{-c}
\end{equation}
here, $z_{3}$ encodes the count of shared questions $N_{\mathrm{Q}}$ between $u$ and $s$. The more questions they both answered, the more similar the two students are. 

\begin{equation}
    z_{4} = (1+N_{\mathrm{K}})^{-c}   
\end{equation}
here, $z_{4}$ encodes the count of shared knowledge concepts $N_{\mathrm{K}}$ between $u$ and $s$. A larger overlap indicates greater similarity.

\begin{equation}
    z_{5} = (1+f)^{-c}
\end{equation}
here, $z_{5}$ encodes the association strength between the candidate student and the target question, based on their meta-path co-occurrence frequency $f$. A higher frequency implies a stronger association.

\noindent\textbf{Mahalanobis‑Driven Similarity Matching.}
Considering the differences in scales and correlations among the five-dimensional similarity features, we measure the similarity between $u$ and $s$ using Mahalanobis distance \cite{de2000mahalanobis}. This metric can automatically normalize feature dimensions and incorporate covariance information. The similarity distance calculation can be defined as:
\begin{equation}
d(u,s)=
\sqrt{\bigl(\mathbf{z}_{u,s}-\boldsymbol{\mu}\bigr)^{\top}
       \boldsymbol{\Sigma}^{-1}
       \bigl(\mathbf{z}_{u,s}-\boldsymbol{\mu}\bigr)}
\label{eq:mahalanobis}
\end{equation}
where \(\boldsymbol{\mu}\) is the mean vector of the feature vectors sampled from a large number of student pairs in the training set,  
and \(\boldsymbol{\Sigma}\) is the corresponding covariance matrix, regularised with Schmidt shrinkage to avoid singularity. Finally, we sort all candidate students in ascending order of $d(u,s)$ and select the Top-$S$ students with the smallest distances as the most similar students to the target student.

\subsection{LLM Prediction Module}
Considering that traditional knowledge tracing models can usually only output numerical indicators and lack high-level semantic explanations of prediction results. In order to address the critical needs of both prediction accuracy and interpretability in educational scenarios, this stage integrates cross‑student interaction information into the prompt and employs the LLM to perform KT prediction and interpretable analysis report generation. Considering that our method employs a zero-shot prediction approach without fine-tuning the LLM, the quality of the prompt template is particularly critical to the prediction. A simple prompt example is shown in Fig.~\ref{fig:Framework}, with the complete version provided in Appendix C.

\noindent\textbf{Structured Prompt Construction.}
The input information consists of three blocks: (1) \textit{target student information}: including student ability, historical interaction sequence, and question attributes (knowledge concept, difficulty, and discrimination parameters obtained from the IRT-2PL model) within the interactions; (2) \textit{target question information}: including question attributes and the student's historical accuracy on the target KC; (3) \textit{similar student information}: including student ability, historical interaction sequences on the target KC and corresponding accuracy.

\noindent\textbf{Prediction and Explanation Generation.}
Once the structured prompt is constructed, we invoke the LLM to perform two tasks in a single pass: (1) \textit{answer result prediction}: including the binary result of the answer $(correct / wrong)$ and the confidence probability; (2) \textit{natural language analysis report}: a three-sentence analysis and explanation report interpreting the prediction process.


\section{Experiment}

In this section, we conduct extensive experiments to illustrate the effectiveness of our proposed method.

\begin{table}[b]
\setlength{\tabcolsep}{1mm}  
\small
\centering
\begin{tabular}{lcccc}
\toprule
\textbf{Datasets} & \textbf{Students} & \textbf{Questions} & \textbf{KCs} & \textbf{Interactions} \\
\midrule
Assistment09 & 3{,}013 & 9{,}795 & 107 & 297{,}575 \\
Slepemapy    & 5{,}000 & 2{,}727 & 1{,}390 & 615{,}042 \\
Statics2011  & 331    & 633   & 97  & 111{,}468 \\
Frcsub       & 536    & 20    & 8   & 98{,}624 \\
\bottomrule
\end{tabular}
\normalsize
\caption{Statistics of the four processed datasets.}
\label{tab:dataset-stats}
\end{table}

\begin{table}[t]
  \centering
  \setlength{\tabcolsep}{1mm}
  \small
  \begin{tabular}{lcccc}
    \toprule
    \multicolumn{1}{l}{\multirow{2}{*}{\textbf{Models}}}
      & \multicolumn{4}{c}{\textbf{AUC}} \\
    \cmidrule(lr){2-5}
      & \textbf{Assistment09} 
      & \textbf{Slepemapy} 
      & \textbf{Statics2011} 
      & \textbf{Frcsub} \\
    \midrule
      DKT      & 0.7452 & 0.7386 & 0.7927 & 0.8695 \\
      DKT+     & 0.7551 & 0.7426 & 0.7995 & 0.8736 \\
      DKT-Forget    & 0.7485 & 0.7448 & 0.7852 & 0.8645 \\ 
      AT‑DKT   & 0.7596 & 0.7551 & 0.8024 & 0.8831 \\
      DKVMN    & 0.7442 & 0.7726 & 0.7774 & 0.9023 \\
      Deep‑IRT & 0.7425 & 0.7584 & 0.7841 & 0.8462 \\
      AKT      & 0.7788 & 0.7952 & 0.8242 & 0.8954 \\
      GKT      & 0.7459 & 0.7010 & 0.7888 & 0.8387 \\
      SAKT     & 0.7133 & 0.7633 & 0.7776 & 0.8469 \\
      CoKT     & \underline{0.8211} & 0.8164 & 0.8270 & \underline{0.9238} \\
    \midrule
      PEBG+DKT & 0.8183 & 0.8213 & 0.8179 & 0.9187 \\
      TCL4KT   & 0.7918 & 0.8018 & \underline{0.8357} & 0.9079 \\
      SimQE    & 0.8027 & 0.8269 & 0.8321 & 0.9163 \\
      STHKT    & 0.8056 & \underline{0.8574} & 0.8295 & 0.9135 \\
    \midrule
      EPFL     & 0.5827 & 0.6088 & 0.7189 & 0.7446 \\
      EFKT     & 0.7945 & 0.6843 & 0.7671 & 0.8758 \\
    \midrule
    \multicolumn{5}{l}{\textit{Ours}} \\
      HISE‑KT\textsubscript{DS} & 0.8381 & 0.9383 & 0.8692 & 0.9322 \\
      HISE‑KT\textsubscript{QP} & \textbf{0.8703} & \textbf{0.9749}
                                & \textbf{0.8888} & \textbf{0.9482} \\
    \bottomrule
  \end{tabular}
  \caption{\textbf{AUC} performance on four datasets. All results are the mean of five runs. Best in \textbf{bold}, second-best \underline{underlined}. \textbf{HISE‑KT\textsubscript{DS}} denotes our method using DeepSeek‑V3; \textbf{HISE‑KT\textsubscript{QP}} denotes our method using Qwen‑Plus.}
  \label{tab: main result auc-only}
\end{table}

\subsection{Experimental Settings}
\noindent\textbf{Datasets.}
We evaluate the performance of HISE-KT on four benchmark datasets: \textbf{Assistment09} \cite{feng2009addressing}, \textbf{Slepemapy} \cite{papouvsek2016adaptive}, \textbf{Statics2011} \cite{koedinger2010data} and \textbf{Frcsub} \cite{wu2015cognitive}. The statistics of the datasets are in Tab.~\ref{tab:dataset-stats}. More details are provided in Appendix D.

\noindent\textbf{Baselines.}
To evaluate the performance of our proposed method, we compare it with three categories of baselines: Deep Learning (DL)-based, HIN-based, and LLM-based methods. The DL-based methods include DKT \cite{piech2015deep}, DKT+ \cite{yeung2018addressing}, DKT-Forget \cite{nagatani2019augmenting}, AT-DKT \cite{liu2023enhancing}, DKVMN \cite{zhang2017dynamic}, Deep-IRT \cite{yeung2019deep}, AKT \cite{ghosh2020context}, GKT \cite{nakagawa2019graph}, SAKT \cite{pandey2019self}, and CoKT \cite{long2022improving}; the HIN-based methods include PEBG+DKT \cite{liu2020improving}, TCL4KT \cite{sun2023weighted}, SimQE \cite{sun2024question}, and STHKT \cite{li2025sthkt}; and the LLM-based methods include EPFL \cite{neshaei2024towards} and EFKT \cite{li2024explainable}. The specific descriptions of each baseline are provided in Appendix E. 



\begin{table}[t]
  \setlength{\tabcolsep}{1mm}  
  \small                        
  \scalebox{1}{
  \centering
  \begin{tabular}{lcccc}
    \toprule
    \multicolumn{1}{l}{\multirow{2}{*}{\textbf{Methods}}}
    & \multicolumn{4}{c}{\textbf{AUC}} \\ 
    \cmidrule(lr){2-5}                   
      & \textbf{Assistment09} 
      & \textbf{Slepemapy} 
      & \textbf{Statics2011} 
      & \textbf{Frcsub} \\
    \midrule
    Full & 0.8703 & 0.9749 & 0.8888 & 0.9482 \\
    w/o MSR   & 0.8593 & 0.9709 & 0.8805 & 0.9399 \\
    w/o MSL   & 0.8473 & 0.9662 & 0.8516 & 0.9321 \\
    w/o SimU  & 0.7734 & 0.6911 & 0.8024 & 0.9177 \\
    w/o RSimU & 0.7754 & 0.6937 & 0.7774 & 0.8692 \\
    w/o IRT   & 0.8492 & 0.9668 & 0.8447 & 0.9248 \\
    \bottomrule
  \end{tabular}
  }
  \normalsize                   
  \caption{Ablation results in terms of \textbf{AUC} on four datasets.}
  \label{tab:ablation}
\end{table}

\noindent\textbf{Implementation Details.}
We divide the dataset into training, validation, and test sets in a ratio of 8:1:1. The LLM used for meta-path scoring is Qwen-Plus \cite{bai2023qwen} and for prediction are Qwen-Plus and DeepSeek-V3 \cite{liu2024deepseek} without additional fine-tuning. For meta-path instantiation, the number of path instances $N$ sampled under each meta-path for each question is 100, and the path length is 20. The scaling constant $c$ in cross-student similarity encoding is set to 2. Parameter selection for Top-\(K\) paths and Top-\(S\) similar students is detailed in the parameter sensitivity study.

\subsection{Main Results}

Tab.~\ref{tab: main result auc-only} shows the AUC performance of all compared models (The complete AUC and ACC results are provided in Appendix F). We find that HISE-KT outperforms all three categories of baseline models, indicating the effectiveness of our proposed HISE-KT method. Compared with DL-based methods, our method improves ACC by 0.74\%-9.68\% and AUC by 2.44\%-15.85\%, showing that it achieves the highest performance without sequence modeling or fine-tuning. Compared with HIN-based methods, our method improves ACC by 1.93\%-7.56\% and AUC by 2.95\%-11.75\%, demonstrating that it selects high-quality meta-paths and extracts more informative structured relationships on MRHIN. Compared with LLM-based methods, our method improves ACC by 1.28\%-24.31\% and AUC by 7.24\%-29.06\%, confirming that integrating cross-student information with LLM reasoning not only yields optimal performance but also enhances interpretability. These results confirm that the synergy between the structured modeling of HIN and the reasoning capabilities of LLM enables accurate and interpretable KT.

\begin{figure}[t]
  \centering
  \begin{subfigure}[b]{\columnwidth}
    \centering
    \includegraphics[width=.24\columnwidth]{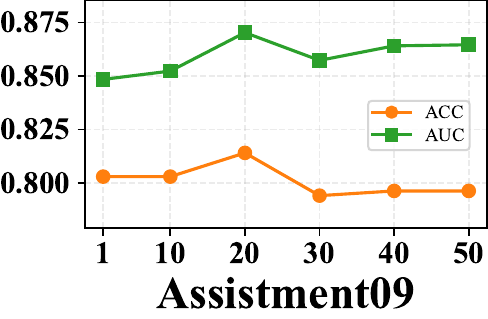}
    \includegraphics[width=.24\columnwidth]{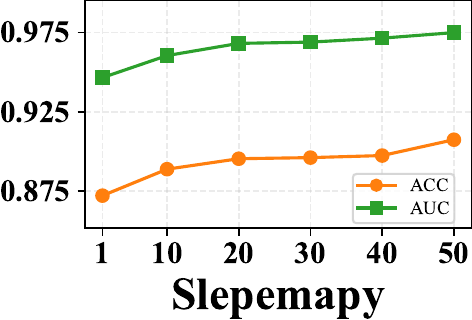}
    \includegraphics[width=.24\columnwidth]{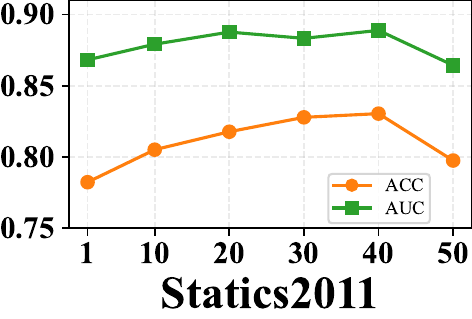}
    \includegraphics[width=.24\columnwidth]{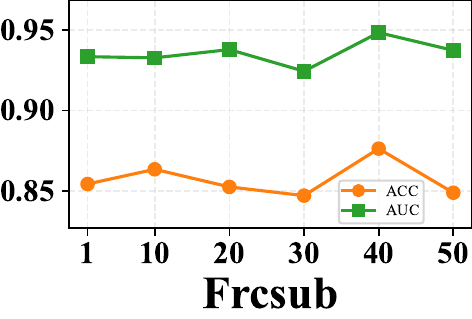}
    \caption{Influence of the number of Top-$K$ meta-path instances.}
    \label{fig:param-topk}
  \end{subfigure}

  
  \begin{subfigure}[b]{\columnwidth}
    \centering
    \includegraphics[width=.24\columnwidth]{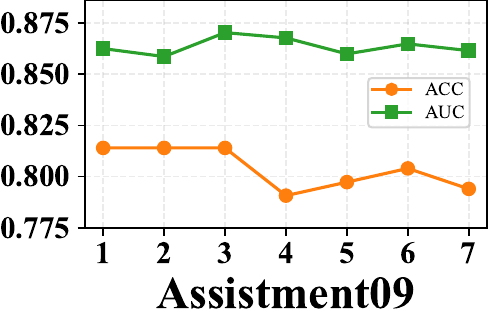}
    \includegraphics[width=.24\columnwidth]{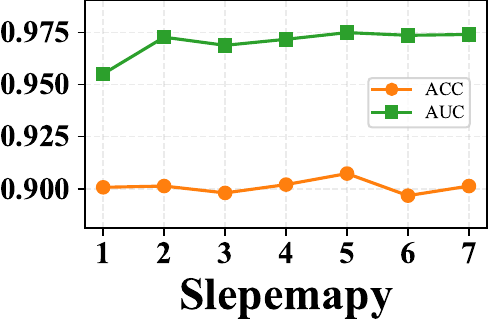}
    \includegraphics[width=.24\columnwidth]{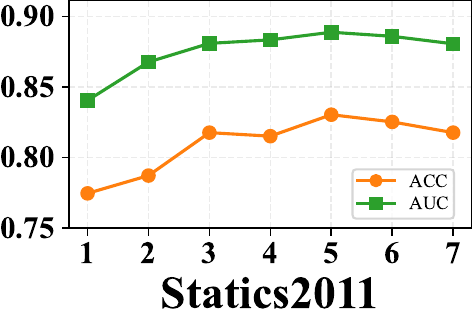}
    \includegraphics[width=.24\columnwidth]{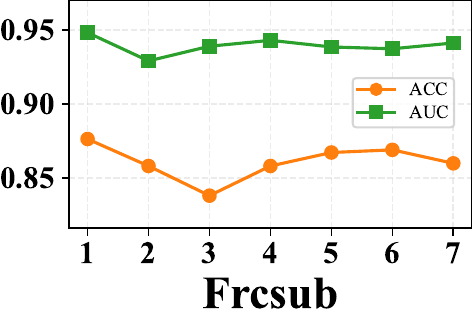}
    \caption{Influence of the number of Top-$S$ similar students.}
    \label{fig:param-similar S}
  \end{subfigure}
  \caption{Parameter‐sensitivity analysis.}
  \label{fig:param-sensitivity}
\end{figure}


\begin{figure*}[t]
  \centering
  \includegraphics[width=0.95\textwidth]{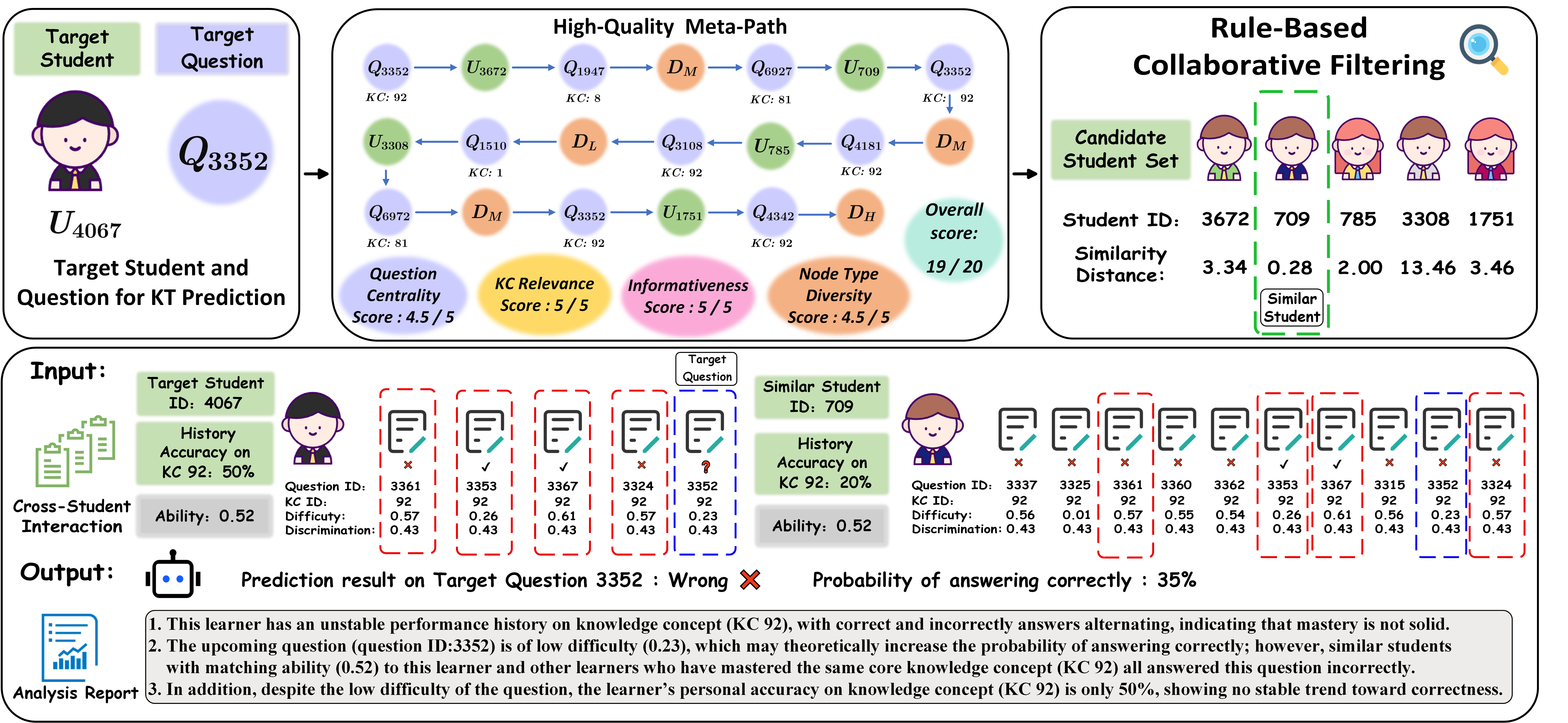}  
  \caption{Case study of a target student $U_{4067}$ on question $Q_{3352}$.}
  \label{fig:case-study}
\end{figure*}

\begin{table*}[t]
  \centering
  \small
  \scalebox{1}{
  \setlength{\tabcolsep}{1mm}
  \begin{tabularx}{\textwidth}{C{0.18\textwidth} L{0.78\textwidth}}
    \toprule
    \textbf{Dimensions} & \textbf{Analysis} \\
    \midrule
     Question Centrality 
      & The target question node \(Q_{3352}\) appears three times and all other question nodes are located close to it. This star-shaped structure shows that the path remains focused on \(Q_{3352}\), yielding a high score. \\
      \midrule
    KC Relevance 
      & Four of the seven questions directly examine $KC_{92}$ while only a few involve other concepts. $KC_{92}$ appears repeatedly, indicating strong conceptual coherence and yielding a high score. \\
      \midrule
    Informativeness 
      & There are no repeated student, non-target question, or knowledge concept nodes. Each step contributes new information without redundancy, yielding a high score. \\
      \midrule
    Node Type Diversity 
      & The path spans three difficulty levels (\(D_{\text{low}}\), \(D_{\text{medium}}\), \(D_{\text{high}}\)) and interleaves them with different question and student nodes, resulting in rich node types and a high diversity score. \\
    \bottomrule
  \end{tabularx}
  }
  \normalsize
  \caption{Analysis of high-quality meta-path instance in the case study.}
  \label{tab:case dimensions analysis}
\end{table*}

\subsection{Ablation Study}
Tab.~\ref{tab:ablation} shows the ablation study results in terms of AUC (The complete AUC and ACC results are provided in Appendix F). To further investigate the contribution of different components in HISE-KT, we design five variants:
    (1) \textbf{w/o MSR} indicates removing the meta-path scorer and using Random-$K$ meta-paths that have not been filtered by the meta-path scorer to replace the Top-$K$ meta-paths; (2) \textbf{w/o MSL} indicates using the Lowest-$K$ scoring meta-paths to replace the Top-$K$ scoring meta-paths; (3) \textbf{w/o SimU} indicates removing information about similar students during LLM prediction; (4) \textbf{w/o RSimU} indicates students randomly selected from the candidate student set are used to replace similar students obtained through similarity retrieval; (5) \textbf{w/o IRT} indicates removing IRT-2PL parameters (e.g., ability, difficulty, discrimination) during KT prediction. 


From Tab.~\ref{tab:ablation}, we can conclude that removing the meta-path scorer leads to a decline in performance, indicating that high-quality meta-paths are crucial for capturing structured relationships in MRHIN. Moreover, the performance using the Lowest-$K$ scoring paths is inferior to that using Random-$K$ paths, further confirming the importance of the meta-path scorer. When similar student information is not incorporated, the performance drops significantly, indicating that the cross-student interaction information is crucial for KT prediction. Additionally, replacing the similar students retrieved by the rule‑based collaborative filtering module with randomly selected ones further degrades performance, suggesting that irrelevant students introduce noise and thereby damage performance.  When the IRT information is removed, the LLM cannot assess whether the difficulty and discrimination of the target question match the target student's ability based on such parameter information, consequently affecting the prediction performance. In summary, all components of HISE-KT contribute
positively to performance, with the meta-path scorer and similar students information being particularly crucial for effective cross-student interaction modeling.

\subsection{Parameter Sensitivity}
Fig.~\ref{fig:param-sensitivity} shows the impact of the number of Top-\(K\) meta-path instances and the number of Top-$S$ similar students on prediction performance. As can be seen from Fig.~\ref{fig:param-topk}, the prediction performance is poor when the Top-\(K\) parameter is selected to be small, indicating that fewer meta-paths cannot mine the rich information between questions and students. When the value of Top-\(K\) increases, the prediction performance gradually improves; however, when Top-\(K\) becomes too large, the model performance tends to show  diminishing returns or slight performance degradation, indicating that relatively low-quality meta-path instances introduce noise that degrades prediction performance.

In Fig.~\ref{fig:param-similar S}, for the Top-\(S\) parameter, different datasets have different optimal numbers of similar students, typically between 3 and 5, whereas the \emph{Frcsub} dataset achieves its best results with Top-$S=1$. Because the questions in this dataset often involve multiple knowledge concepts, introducing too many similar students shows inconsistent mastery of multiple knowledge concepts, which tends to interfere with the LLM’s judgment. These results demonstrate the relative robustness of HISE-KT to hyperparameter configurations within reasonable ranges.

\subsection{Case Study}


We select a high-quality meta-path instance to demonstrate the entire prediction process. Fig.~\ref{fig:case-study} shows the process, and Tab.~\ref{tab:case dimensions analysis} lists the quality analysis results of the meta-path instance. First, given the target student \(U_{4067}\) and the target question \(Q_{3352}\), we extract the high-quality meta-path instance corresponding to the question. Then, based on this path, we retrieve the most similar student \(U_{709}\), and input the interaction history between \(U_{4067}\) and \(U_{709}\) into the LLM to complete the prediction of \(U_{4067}\)'s performance and generate a three-sentence explainable analysis report. It is noteworthy that
\(U_{4067}\)’s accuracy on \(KC_{92}\) is 50\% with 
significant performance fluctuations. It is difficult to make a reliable judgment solely based on \(U_{4067}\)'s own data. Meanwhile, \(U_{709}\)’s 
interaction history completely overlaps with \(U_{4067}\), and the answer result of \(Q_{3352}\) is wrong. Based on the cross-student information, the LLM predicts that \(U_{4067}\) also answers incorrectly.

\section{Conclusion}
In this paper, we propose an innovative framework that synergistically integrates heterogeneous information networks with large language models called HISE-KT to achieve accurate and explainable knowledge tracing. Key innovations include deploying LLMs for automated meta-path quality assessment to eliminate noise, designing an educational psychology-inspired mechanism to retrieve similar student trajectories for enriched context, and leveraging structured prompts to fuse historical data with relational evidence. This integration enables simultaneous prediction and evidence-based explanation generation. Extensive experiments across four public benchmarks demonstrate HISE-KT’s superior accuracy and unprecedented interpretability over state-of-the-art baselines. Our research establishes a new paradigm for developing transparent, adaptive educational AI through principled fusion of structural and semantic modeling.



\section{Acknowledgments}
This work was funded by the National Natural Science Foundation of China (Nos. 62567005 and 62206107), and Natural Science Foundation of Inner Mongolia Autonomous Region of China (No. 2025MS06004).

\bibliography{main}



\clearpage
\begin{appendices}
\renewcommand\thesection{\Alph{section}}     
\setcounter{secnumdepth}{2}   

\appendix

\section{External Node Construction}
\subsection{IRT-2PL Formula}
We employ the two-parameter logistic Item Response Theory (IRT-2PL) model to estimate student ability and question difficulty parameters to construct the two external nodes $A$ and $D$. Formally, the probability that student \(i\) answers question \(j\) correctly is formulated as: 

\begin{equation}
P(c_{ij}=1\mid\theta_i,b_j,a_j)=\frac{1}{1+\exp\ \!\bigl(-a_j(\theta_i-b_j)\bigr)}
\end{equation}
where \(\theta_i\) is the latent ability parameter of student \(i\), \(b_j\) is the difficulty parameter of question \(j\), and \(a_j\) is the discrimination parameter of question \(j\).

\subsection{Level Division Formula}
Since continuous numerical parameters are not suitable to be used directly as nodes in MRHIN. Based on the normal distribution division method, we discretize student ability \(\theta_i\) and question difficulty \(b_j\), and use the mean \(\mu\) and standard deviation \(\sigma\) of their values to divide student ability and question difficulty into three levels: Low, Medium, and High, which are formulated as:

\begin{equation}
\text{level}(x)=
\left\{
\begin{aligned}
\text{Low},    &\quad x < \mu - \sigma\\
\text{Medium}, &\quad \mu - \sigma \le x \le \mu + \sigma\\
\text{High},   &\quad x > \mu + \sigma
\end{aligned}
\right.
\end{equation}
where \(x\) denotes either a student ability parameter \(\theta_i\) or a question difficulty parameter \(b_j\).

\section{Composite Meta-Path}
\textit{Composite meta-paths} are composed of basic paths. In order to capture the complex cross-semantic relationships between multiple nodes, we define ten meta-paths.
\begin{itemize}
    \item \textit{Q-K-Q-D-Q} and \textit{Q-D-Q-K-Q}: the two meta-paths capture the relationship among $Q$, $K$, and $D$.
    \item \textit{Q-U-Q-D-Q} and \textit{Q-D-Q-U-Q}: the two meta-paths capture the relationship among $Q$, $U$, and $D$.
    \item \textit{Q-K-Q-U-Q} and \textit{Q-U-Q-K-Q}: the two meta-paths capture the relationship among $Q$, $K$, and $U$.    
    \item \textit{Q-K-Q-U-Q-D-Q} and \textit{Q-U-Q-K-Q-D-Q}: the two meta-paths capture the relationship among $Q$, $K$, $U$, and $D$.    
    \item \textit{Q-K-Q-U-A-U-Q}: this meta-path captures the relationship among $Q$, $K$, $U$, and $A$.
    \item \textit{Q-K-Q-U-Q-D-Q-U-A-U-Q}: this meta-path captures the relationship among all five node types $Q$, $K$, $U$, $A$, and $D$.
\end{itemize}
\label{app:Composite Meta-Path}

\section{Full Prompt Templates}
\subsection{Meta-Path Scoring}
\begin{figure*}[t]
    \centering
    \includegraphics[width=\textwidth]{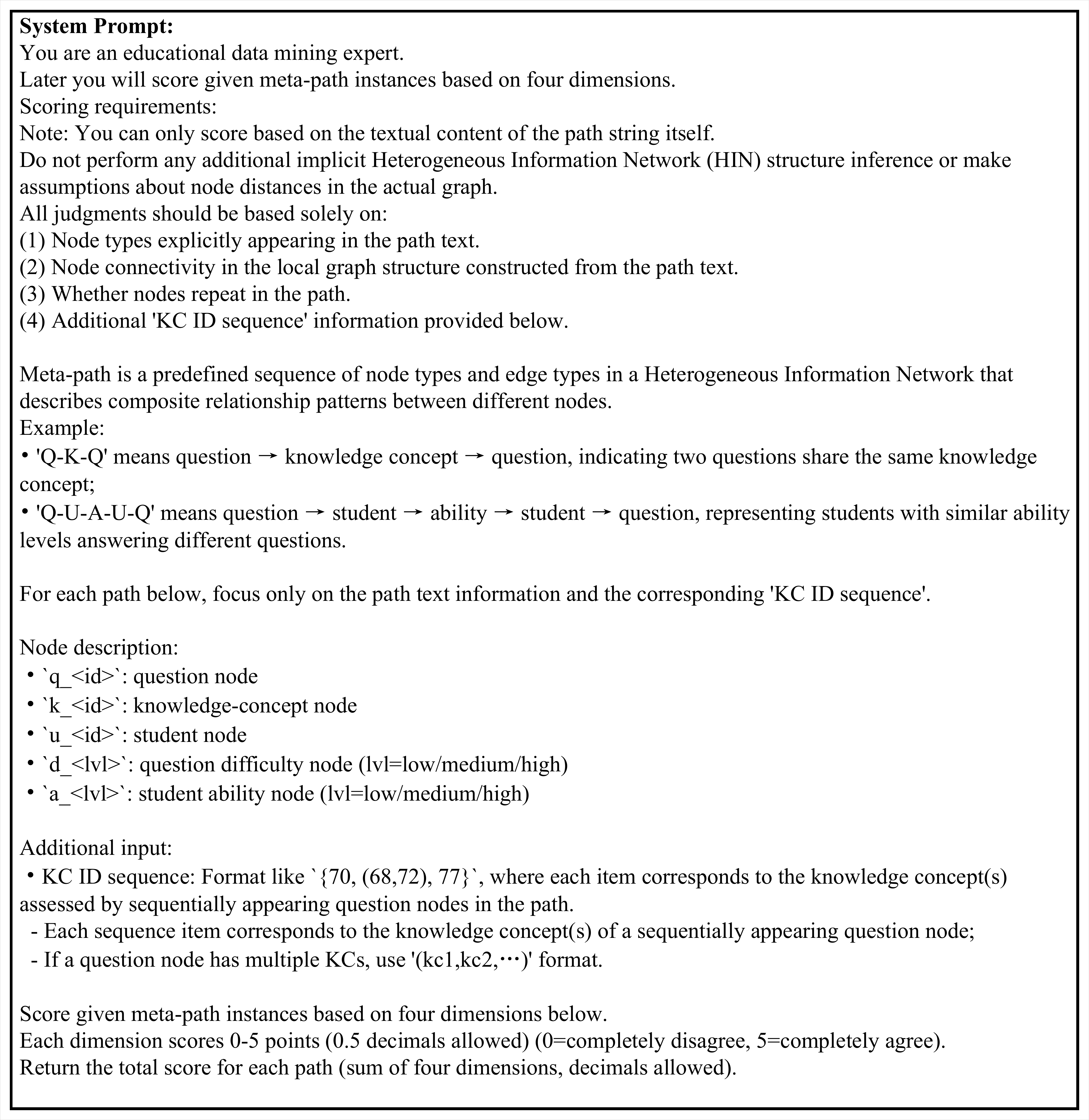}
    \caption{Prompt of meta-path scoring (1)}
    \label{fig:Meta-Path Scoring (1)}
\end{figure*}

\begin{figure*}[t]
    \centering
    \includegraphics[width=\textwidth]{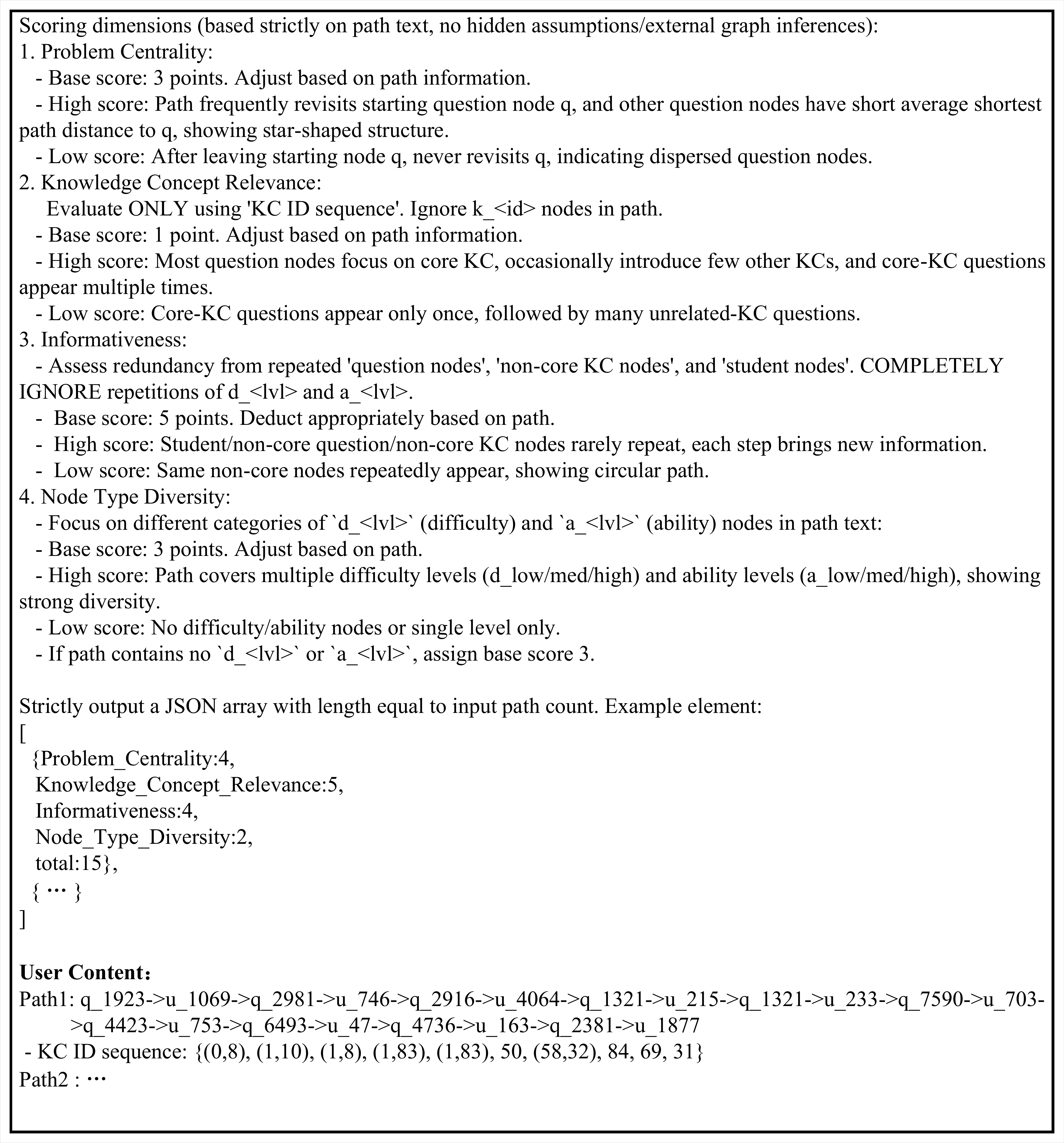}
    \caption{Prompt of meta-path scoring (2)}
    \label{fig:Meta-Path Scoring (2)}
\end{figure*}

The meta-path scoring prompt is shown in Fig.~\ref{fig:Meta-Path Scoring (1)} and Fig.~\ref{fig:Meta-Path Scoring (2)} at the end of the document.

\subsection{LLM Prediction}
\begin{figure*}[t]
    \centering
    \includegraphics[width=\textwidth]{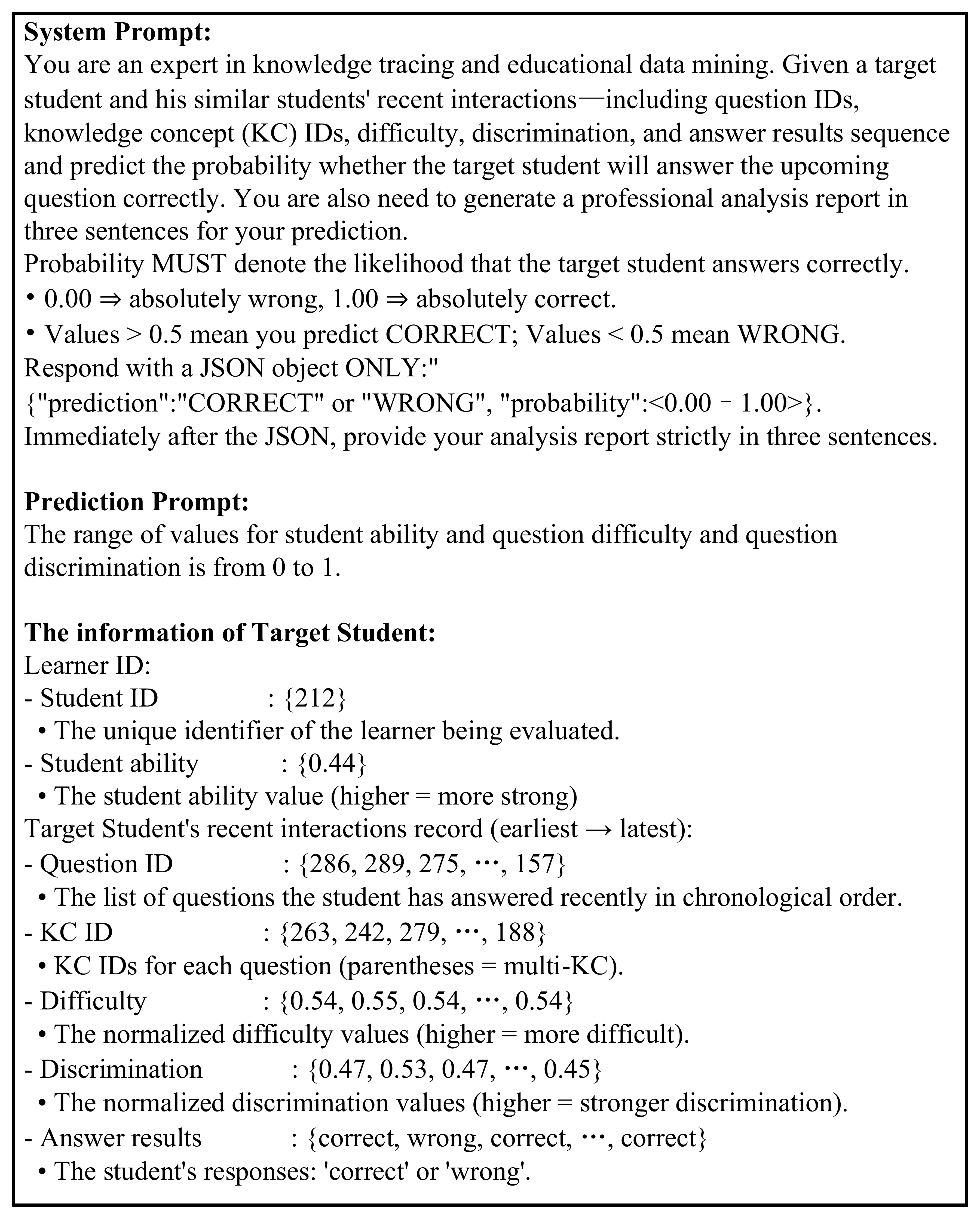}
    \caption{Prompt of LLM prediction (1)}
    \label{fig:LLM Prediction (1)}
\end{figure*}

\begin{figure*}[t]
    \centering
    \includegraphics[width=\textwidth]{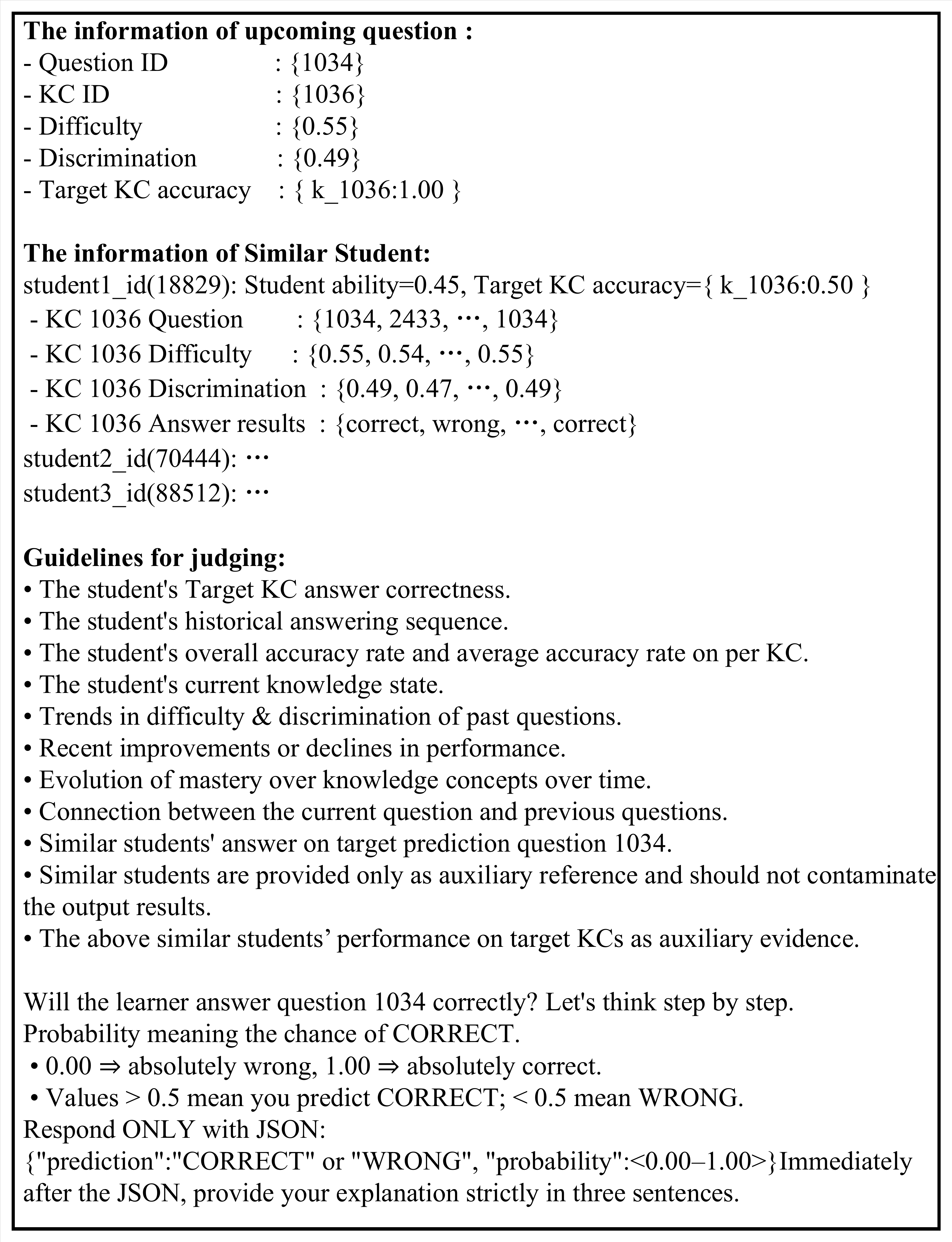}
    \caption{Prompt of LLM prediction (2)}
    \label{fig:LLM Prediction (2)}
\end{figure*}

The LLM prediction prompt is shown in Fig.~\ref{fig:LLM Prediction (1)} and Fig.~\ref{fig:LLM Prediction (2)} at the end of the document.

\label{app:prompt}

\section{Datasets}
\par We evaluate the performance of HISE-KT on four commonly used public datasets: 

\begin{itemize}
\item \noindent\textbf{Assistment09\footnote{%
\url{https://sites.google.com/site/assistmentsdata/home/2009-2010-assistment-data/skill-builder-data-2009-2010}%
}} is collected from the ASSISTments online learning platform during the 2009-2010 academic year; the questions focus primarily on mathematics.

\item \noindent\textbf{Slepemapy\footnote{%
\url{https://www.fi.muni.cz/adaptivelearning/data/slepemapy/}%
}} is sourced from the Czech geography practice site \emph{Slepemapy.cz}, where students label countries, cities, or landforms on blank maps; the items target geography skills.

\item \noindent\textbf{Statics2011\footnote{%
\url{https://pslcdatashop.web.cmu.edu/DatasetInfo?datasetId=507}%
}} is collected from an Engineering Statics course taught at Carnegie Mellon University in Fall 2011.

\item \noindent\textbf{Frcsub\footnote{%
\url{https://base.ustc.edu.cn/data/math2015.rar}%
}} mainly focuses on the middle school students’ responses to fraction subtraction questions.
\end{itemize}
we remove students with less than 10 interaction records and questions that are answered less than 10 times. For the Slepemapy dataset, we randomly select the interaction records of 5{,}000 students in our experiments. 
\label{app:dataset}

\begin{table*}[t]
\centering
\scalebox{1}{%
\setlength{\tabcolsep}{1mm}  
\begin{tabular}{cccccccccc}
\toprule
\multicolumn{1}{c}{\multirow{2}{*}{\textbf{Categories}}} &
\multicolumn{1}{c}{\multirow{2}{*}{\textbf{Models}}} &
\multicolumn{2}{c}{\textbf{Assistment09}} &
\multicolumn{2}{c}{\textbf{Slepemapy}} &
\multicolumn{2}{c}{\textbf{Statics2011}} &
\multicolumn{2}{c}{\textbf{Frcsub}} \\
\cmidrule(lr){3-4}\cmidrule(lr){5-6}\cmidrule(lr){7-8}\cmidrule(lr){9-10}
\multicolumn{1}{c}{} & \multicolumn{1}{c}{} & ACC & AUC & ACC & AUC & ACC & AUC & ACC & AUC \\
\midrule
\multirow{10}{*}{\textit{DL-based Methods}}
 & DKT              & 0.7218 & 0.7452 & 0.7824 & 0.7386 & 0.7915 & 0.7927 & 0.7952 & 0.8695 \\
 & DKT+             & 0.7258 & 0.7551 & 0.7973 & 0.7426 & 0.8012 & 0.7995 & 0.7964 & 0.8736 \\
 & DKT-Forget  & 0.7216 & 0.7485 & 0.7822 & 0.7448 & 0.7926 & 0.7852 & 0.7913 & 0.8645 \\
 & AT-DKT           & 0.7263 & 0.7596 & 0.7934 & 0.7551 & 0.7988 & 0.8024 & 0.8017 & 0.8831 \\
 & DKVMN            & 0.7129 & 0.7442 & 0.7768 & 0.7726 & 0.7668 & 0.7774 & 0.8287 & 0.9023 \\
 & Deep-IRT         & 0.7218 & 0.7425 & 0.7818 & 0.7584 & 0.8048 & 0.7841 & 0.7839 & 0.8462 \\
 & AKT              & 0.7380 & 0.7788 & 0.8065 & 0.7952 & 0.8230 & 0.8242 & 0.8218 & 0.8954 \\
 & GKT              & 0.7224 & 0.7459 & 0.7962 & 0.7010 & 0.7862 & 0.7888 & 0.7589 & 0.8387 \\
 & SAKT             & 0.6976 & 0.7133 & 0.7941 & 0.7633 & 0.7749 & 0.7776 & 0.7947 & 0.8469 \\
 & CoKT             & \underline{0.7709} & \underline{0.8211} & 0.8105 & 0.8164 & 0.8081 & 0.8270 & 0.8432 & \underline{0.9238} \\
\midrule
\multirow{4}{*}{\textit{HIN-based Methods}}
 & PEBG+DKT         & 0.7668 & 0.8183 & 0.8154 & 0.8213 & 0.8013 & 0.8179 & 0.8246 & 0.9187 \\
 & TCL4KT           & 0.7561 & 0.7918 & 0.7937 & 0.8018 & 0.8092 & \underline{0.8357} & 0.8187 & 0.9079 \\
 & SimQE            & 0.7643 & 0.8027 & 0.8173 & 0.8269 & \underline{0.8111} & 0.8321 & 0.8255 & 0.9163 \\
 & STHKT            & 0.7682 & 0.8056 & \underline{0.8317} & \underline{0.8574} & 0.8082 & 0.8295 & 0.8212 & 0.9135 \\
\midrule
\multirow{2}{*}{\textit{LLM-based Methods}}
 &  EPFL             & 0.6013 & 0.5827 & 0.6160 & 0.6088 & 0.6962 & 0.7189 & \underline{0.8636} & 0.7446 \\
 &  EFKT             & 0.7641 & 0.7945 & 0.6660 & 0.6843 & 0.7342 & 0.7671 & 0.8091 & 0.8758 \\
\midrule
\multirow{2}{*}{\textit{Ours}}
 & HISE-KT\textsubscript{\text{DeepSeek-V3}} & 0.7993 & 0.8381 & 0.8680 & 0.9383 & 0.8076 & 0.8692 & 0.8764 & 0.9322 \\
 & HISE-KT\textsubscript{\text{Qwen-Plus}}   & \textbf{0.8140} & \textbf{0.8703} & \textbf{0.9073} & \textbf{0.9749} & \textbf{0.8304} & \textbf{0.8888} & \textbf{0.8764} & \textbf{0.9482} \\
\bottomrule
\end{tabular}%
\normalsize                   
}
\caption{Prediction performance on four datasets. The best results are shown in bold, and the second best results are underlined.}
\label{tab:main results AUC and ACC}
\end{table*}

\begin{table*}[t]
\centering
\scalebox{1}{
\begin{tabular}{lcccccccc}
\toprule
\multirow{2}{*}{\textbf{Methods}} &
\multicolumn{2}{c}{\textbf{Assistment09}} &
\multicolumn{2}{c}{\textbf{Slepemapy}} &
\multicolumn{2}{c}{\textbf{Statics2011}} &
\multicolumn{2}{c}{\textbf{Frcsub}} \\
\cmidrule(lr){2-3}\cmidrule(lr){4-5}\cmidrule(lr){6-7}\cmidrule(lr){8-9}
& ACC & AUC & ACC & AUC & ACC & AUC & ACC & AUC \\
\midrule
Full              & 0.8140 & 0.8703 & 0.9073 & 0.9749 & 0.8304 & 0.8888 & 0.8764 & 0.9482 \\
w/o MSR           & 0.7920 & 0.8593 & 0.8973 & 0.9709 & 0.8076 & 0.8805 & 0.8545 & 0.9399 \\
w/o MSL           & 0.7841 & 0.8473 & 0.8960 & 0.9662 & 0.8025 & 0.8516 & 0.8491 & 0.9321 \\
w/o SimU          & 0.7176 & 0.7734 & 0.6620 & 0.6911 & 0.7656 & 0.8024 & 0.8436 & 0.9177 \\
w/o RSimU         & 0.7575 & 0.7754 & 0.6407 & 0.6937 & 0.7245 & 0.7774 & 0.8000 & 0.8692 \\
w/o IRT           & 0.7896 & 0.8492 & 0.8987 & 0.9668 & 0.7975 & 0.8447 & 0.8564 & 0.9248 \\

\bottomrule
\end{tabular}
}
\normalsize
\caption{Ablation study on four datasets.}
\label{tab:ablation AUC and ACC}
\end{table*}

\section{Baselines}
To evaluate the performance of our proposed method, we compare it with three categories of baselines: DL-based, HIN-based, and LLM-based methods.

\subsection{DL-based methods}
The DL-based methods include:
\begin{itemize}
    \item \textbf{DKT} uses recurrent neural networks (RNNs) to encode the student's answer sequence.
    \item \textbf{DKT+} introduces regularization terms corresponding to reconstruction error and fluctuation metric in its loss function to improve DKT.
    \item \textbf{DKT-Forget} considers students' forgetting behavior to improve DKT.
    \item \textbf{AT-DKT} improves DKT by introducing two auxiliary learning tasks.
    \item \textbf{DKVMN} uses a dynamic key-value memory network to trace students' knowledge state.
    \item \textbf{Deep-IRT} enhances the interpretability of the model by combining IRT and DKVMN.
    \item \textbf{AKT} uses the Rasch model to obtain a series of Rasch model-based embeddings to capture individual differences in responses.
    \item \textbf{GKT} employs a graph structure to  propagate students’ knowledge states.
    \item \textbf{SAKT} employs self-attention mechanism to capture the relation between interactions.
    \item \textbf{CoKT} considers collaborative information among students to improve KT.
\end{itemize}

\subsection{HIN-based methods}
The HIN-based methods include:
\begin{itemize}
    \item \textbf{PEBG+DKT} uses a bipartite question–skill graph and a product layer to learn low-dimensional question embeddings that capture side information.
    \item \textbf{TCL4KT} applies three-view contrastive learning on a weighted heterogeneous graph to learn enriched question embeddings.
    \item \textbf{SimQE} captures the similarity of questions through biased random walks of metapaths on a weighted HIN to enhance question representation.
    \item \textbf{STHKT} constructs heterogeneous graphs and combines topological Hawkes processes with graph convolutional networks to fuse spatiotemporal information.
\end{itemize}

\subsection{LLM-based methods}
The LLM-based methods include:
\begin{itemize}
    \item \textbf{EPFL} applies pre-trained LLMs via zero-shot prompting and fine-tuning to model student learning trajectories.
    \item \textbf{EFKT} utilizes few-shot prompting to leverage the reasoning and generation capabilities of LLMs for KT task.
\end{itemize}

\label{app:baselines}

\section{Additional Experimental Results}
\subsection{Main Result}
Tab.~\ref{tab:main results AUC and ACC} shows the complete prediction performance of all compared models (both AUC and ACC).  


\subsection{Ablation Study}
Tab.~\ref{tab:ablation AUC and ACC} shows the complete ablation study results (both AUC and ACC).

\label{app:exp}








\end{appendices}

\end{document}